\documentclass[runningheads]{llncs}
\usepackage{graphicx}
\usepackage{cite}
\usepackage{color}
\usepackage{multirow}
\usepackage{url}
\usepackage[misc]{ifsym} 
\usepackage{tablefootnote}

\begin{document}
\title{Not End-to-End: Explore Multi-Stage Architecture for Online Surgical Phase Recognition}

\author{Fangqiu Yi 
	\and
	{Tingting Jiang} 
}

\authorrunning{F. Yi et al.}
%
\institute{NELVT, Department of Computer Science, Peking University, China
	\email{\{chinayi,ttjiang\}@pku.edu.cn}}
\maketitle              
\begin{abstract}
Surgical phase recognition is of particular interest to computer assisted surgery systems, in which the goal is to predict what phase is occurring at each frame for a surgery video. Networks with multi-stage architecture have been widely applied in many computer vision tasks with rich patterns, where a predictor stage first outputs initial predictions and an additional refinement stage operates on the initial predictions to perform further refinement. Existing works show that surgical video contents are well ordered and contain rich temporal patterns, making the multi-stage architecture well suited for the surgical phase recognition task. However, we observe that when simply applying the multi-stage architecture to the surgical phase recognition task, the end-to-end training manner will make the refinement ability fall short of its wishes. To address the problem, we propose a new non end-to-end training strategy and explore different designs of multi-stage architecture for surgical phase recognition task. For the non end-to-end training strategy, the refinement stage is trained separately with proposed two types of disturbed sequences. Meanwhile, we evaluate three different choices of refinement models to show that our analysis and solution are robust to the choices of specific multi-stage models. We conduct experiments on two public benchmarks, the M2CAI16 Workflow Challenge, and the Cholec80 dataset. Results show that multi-stage architecture trained with our strategy largely boosts the performance of the current state-of-the-art single-stage model. Code is avaliable at \url{https://github.com/ChinaYi/casual_tcn}.

\end{abstract}
\section{Introduction}
Surgical phase recognition is of particular interest to computer assisted surgery systems, because it offers solutions to numerous demands of the modern operating room, such as monitoring surgical processes~\cite{BRICONSOUF20072}, scheduling surgeons~\cite{Bhatia:2007:RIO:1620113.1620126} and enhancing coordination among surgical teams~\cite{Lin}. This paper works on the online surgical phase recognition task, which requires to predict what phase is occurring at each frame without using the information of future frames.

Existing models can be divided into two groups. The first group is single-stage models which output prediction results with the input visual features. For example, a number of works utilize dynamic time warping~\cite{Blum2010,Padoy2012Statistical}, conditional random field~\cite{Tao2013Surgical}, and variations of Hidden Markov Model(HMM)~\cite{Lalys2011Surgical,Padoy2008On} over extracted visual features. Jin et al.~\cite{jin2018sv} train an end-to-end RNN model that first uses a very deep ResNet to extract visual features for each frame and then applies a LSTM network to model the temporal dependencies of sequential frames. The second group is the multi-stage models which additionally stacks a refinement stage over the prediction results to perform a further refinement. Czempiel et al.~\cite{tecno} first bring in multi-stage architecture for surgical phase recognition task. They use a causal TCN~\cite{ED-TCN} to output initial predictions over pre-extracted CNN features, and then append another causal TCN~\cite{ED-TCN} to refine the predictions. 

\begin{figure}[t]
	\includegraphics[width=\textwidth]{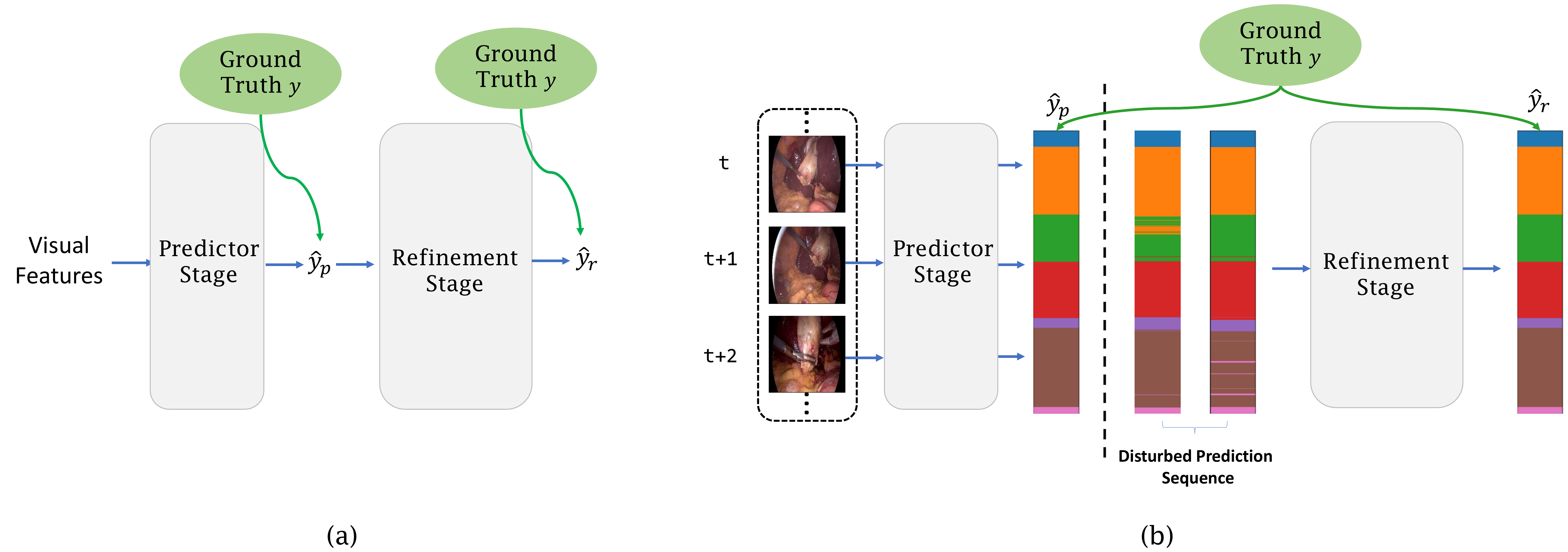}
	\caption{(a). Pipeline of Multi-stage architecture. (b). Our non end-to-end training strategy where two stages are trained separately.} \label{intro_pic}

\end{figure}

In our opinion, the multi-stage architecture is well-suited for the surgical phase recognition task. First of all, networks with multi-stage architecture have been widely applied in many computer vision tasks with rich patterns, such as human pose estimation~\cite{pose_ms1,pose_ms2} and action segmentation~\cite{ms-tcn}. Generally speaking, the idea of multi-stage architecture consists of a predictor stage and a refinement stage, as shown in Fig.~\ref{intro_pic}(a). Sometimes, due to the hard-to-recognize visual features, the initial predictions output by the predictor stage may have errors that violate intrinsic patterns within the data.(\textit{i.e.} The tiny spikes of over-segmentation errors for a continuous action or human pose estimation results that do not conform to the connections of human body joint.) The initial predictions $\hat{y}_p$ are thus further refined by the refinement stage. To avoid the disturbance of noisy visual features, the refinement stage \textbf{only} takes the initial predictions as the input, which forces the refinement process focus on the data intrinsic patterns. Secondly, surgical video contents are well ordered and contain rich temporal patterns~\cite{jin2018sv,hard_frame}. Some works have been motivated by utilizing the rich temporal patterns to refine the predictions. For example, Jin et al.~\cite{jin2018sv} summarize the phase transition conditions(\textit{i.e.} \textit{GallbladderRetraction} cannot happen before \textit{GallbladderPackaging} triggers in Cholec80~\cite{Twinanda_endo} dataset) and propose a post-processing scheme to calibrate the predictions that do not satisfy the transition priors. Yi et al.~\cite{hard_frame} propose a mapping model to decide the phase label of the detected hard frames according to the previous predictions. Their success show that it is possible for the multi-stage architecture to rectify the misclassifications due to the ambiguous visual features in the predictor stage. 

However, we observe that the improvement of multi-stage structure in the surgical phase recognition task is not as obvious as other tasks. Experiments in Czempiel et al.\cite{tecno} show that the performance improvement of the additional refinement stage is very limited. This interesting phenomenon raises our concerns, \textbf{why} the multi-stage architecture does not work as well as we expected? 

In this paper, we first answer the question of \textbf{why} with our analysis, and then further give a solution of \textbf{how} to make multi-stage architecture work better for surgical phase recognition task. The reason of \textbf{why} are two folds. Firstly, a common issue for the multi-stage architecture is that the refinement stage cannot actually learn how to refine with an end-to-end training manner. As shown in Fig.~\ref{intro_pic}(a), the supervision signal is applied in both the predictor stage and refinement stage. After several epochs of training, the predictor stage will quickly converge to the ground truth, which means that the initial predictions $\hat{y}_p$ are almost 100\% correct during the training process. However, in the inference process with the test data, $\hat{y}_p$ still remain lots of errors due to imperfect predictor. The huge gap between the inputs of the refinement stage during training and inference makes the refinement ability fall short of its wishes. Secondly, in the case of end-to-end training, the limited size of current datasets for surgery phase recognition cannot afford the training of refinement stage which brings additional parameters. This leads to a severe overfitting problem compared to the results of applying multi-stage architecture to other vision tasks. 

With the answer of \textbf{why}, we propose a non end-to-end training strategy where the predictor stage and the refinement stage are trained separately to solve the above two issues simultaneously. As shown in Fig.~\ref{intro_pic}(b), the predictor stage is trained with the raw video data. To reduce the gap between the inputs of the refinement stage during training and inference, two types of training sequences for refinement stage are carefully designed to simulate the real output of the predictor during inference, denoted as cross-validate type and mask-hard-frame type respectively. Meanwhile, as the refinement stage is separately trained with the two types of carefully designed training sequences, the over-fitting problem for surgical phase recognition can also be alleviated.

Besides the training strategy, we further explore the designs of multi-stage architecture by evaluating three different temporal models for the refinement stage, including TCN~\cite{ED-TCN}(offline), causal TCN~\cite{ED-TCN}(online) and GRU~\cite{gru}(online). As for the predictor stage, we use the causal TCN in Czempiel et al.\cite{tecno} for its high efficiency and SOTA performance. In principle, our solution can be applied to any single-stage predictor model. Extensive experiments are conducted on two public benchmarks, M2CAI16~\cite{StauderOKKFN16} and Cholec80~\cite{Twinanda_endo}. Results show that all three refinement models trained with our strategy successfully boost the performance of the predictor stage, demonstrating that our analysis and solution are robust to different choices of refinement models. And the SOTA results shows that the multi-stage architecture holds the great potential to boost the performance of existing single-stage models.

\section{Methods}
The multi-stage architecture stacks a refinement stage over the predictor stage sequentially. We propose a training strategy where these two stages are trained separately and explore designs of multi-stage architecture. We first introduce the predictor stage and its training in Sec.~\ref{ps}, then discuss the refinement stage and its training in Sec.~\ref{rs}. It is worth noting that, although we train the multi-stage architecture in a non end-to-end manner, the inference process is still end-to-end as the normal multi-stage architecture. 

\subsection{Predictor Stage}
\label{ps}
We use causal TCN in Czempiel et al.\cite{tecno} to get the initial predictions for its high efficiency and SOTA performance. In principle, the predictor model can be any single-stage models. The input of the causal TCN is the frame-wise extracted features from a pre-trained CNN. Denote the output prediction sequence as $\hat{y}_p \in \mathcal{R}^{C \times T}$. For each frame, the output is a vector of size $C$ denoting the classification probability for each class. For the loss function $\mathcal{L}_p$, we use a combination of cross-entropy classification loss and a smoothing loss~\cite{MSTCN} by deploying a mean square error over the classification probabilities of every two adjacent frames.

\begin{equation}
\mathcal{L}_{p} = \frac{1}{T} \sum_{t=1}^{T}-log(\hat{y}_{p(c,t)}) + \frac{1}{TC} \sum_{m=1}^{C}\sum_{t=1}^{T-1}{|\hat{y}_{p(m,t)}-\hat{y}_{p(m,t+1)}|}^2
\end{equation}

\subsection{Refinement Stage}
\label{rs}
The input of the refinement stage is the prediction results $\hat{y}_p$ output by the predictor stage. During the inference, $\hat{y}_p$ will remain lots of errors due to the imperfect predictor. In order to achieve better refinement results, we should minimize the distribution gap between the training and test. Thus, we design two types of disturbed prediction sequences by simulating the imperfect predictions of the predictor model during the inference. Note that the generation of both two types of disturbed prediction sequences are related to the predictor model, since we cannot directly obtain the prediction sequences from the raw video data.

\begin{figure}[t]
	\includegraphics[width=\textwidth]{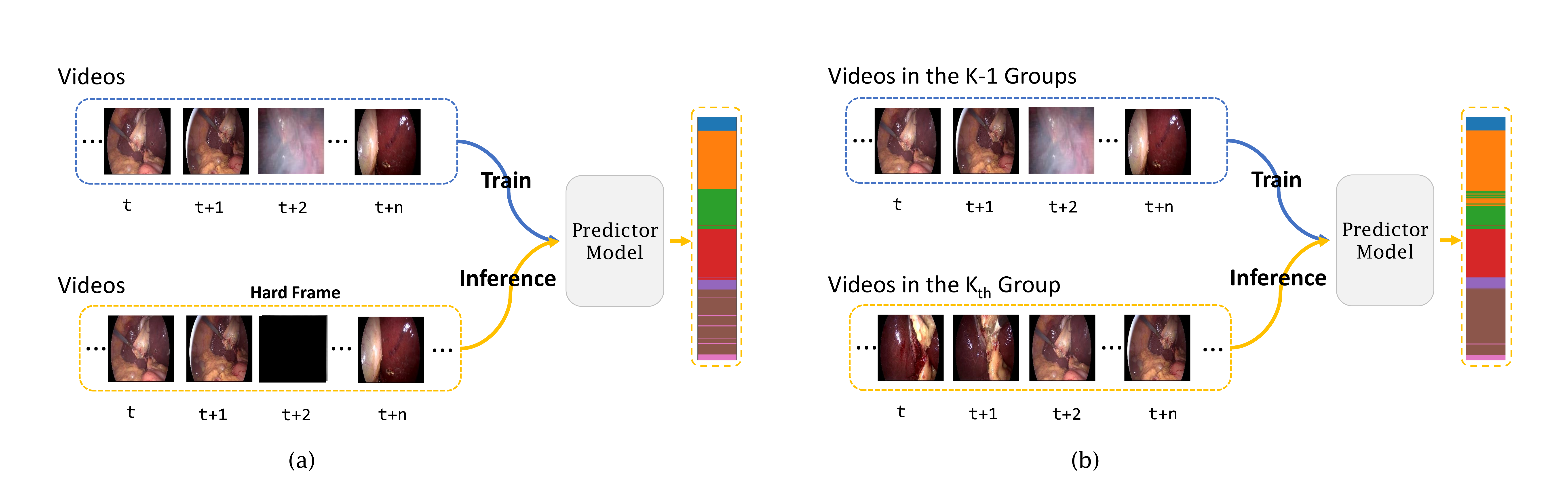}
	\caption{(a). The generation of mask-hard-frame type. (b). The generation of cross-validate type.} \label{two_type_sequence}

\end{figure}

\noindent
\textbf{Mask-Hard-Frame Type.} The concept of hard frames is proposed by Yi et al.~\cite{hard_frame}, which denotes the frames that are not recognizable from their visual appearance. Their work shows that single-stage models usually make mistakes on hard frames. Motivated by this, we seek to add perturbations to the prediction of these hard frames. We first train a predictor model with the normal video data. Then, we follow the rule of Yi et al.~\cite{hard_frame} to find out hard frames in the training set and add perturbations on those hard frames by using a black mask to cover the whole images. Finally, we pass the perturbed video data to the predictor model to get the disturbed prediction sequence. The workflow is shown in Fig.~\ref{two_type_sequence}(a). 

\noindent
\textbf{Cross-Validate Type.} Cross Validate is the simplest way to make part of training data unseen to the predictor model, so that the part of unseen data could be used to simulate the real predictions of the predictor model during the inference. Specifically, we randomly partition the training videos into K groups of equal size. Each time, a single group is retained for validation, and the remaining K-1 groups are used to train a predictor model. And then, we use the trained predictor model to obtain the predictions of the retained video to get the prediction sequence. The workflow is shown in Fig.~\ref{two_type_sequence}(b). We set K = 10 in our experiment. 

For a training set with $N$ training videos, both two types can generate $N$ perturbed prediction sequence to train the refinement model. Both two types of disturbed sequences are used to train the refinement stage in following experiments except for special specification. For the loss function of the refinement stage, we use the cross-entropy loss. For the choice of specific refinement model, we evaluate three common temporal models, including TCN, causal TCN and GRU. The network architectures can be found in the supplementary material.

\section{Experiement}
\subsection{Dataset}
\noindent
\textbf{M2CAI16 Workflow Challenge} dataset~\cite{StauderOKKFN16} contains 41 laparoscopic videos that are acquired at 25fps of cholecystectomy procedures, and 27 of them are used for training and 14 videos are used for testing. These videos are segmented into 8 phases by experienced surgeons.

\noindent
\textbf{Cholec80} dataset~\cite{Twinanda_endo} contains 80 videos of cholecystectomy surgeries performed by 13 surgeons. The dataset is divided into training set (40 videos) and testing set (40 videos). The videos are divided into 7 phases and are captured at 25 fps.

We use three common metrics~\cite{jin2018sv} including the jaccard index (\textit{JACC}), Recall (\textit{Rec}) and Accuracy (\textit{Acc}). The detail training settings can be found in the supplementary material, and code will be published once upon acceptance. 


\begin{table}[t]
	\centering
	\caption{Comparison of multi-stage architecture with three different choices of refinement models under the end-to-end training and ours on Cholec80 dataset. \textit{NULL} denotes single-stage predictor without the refinement stage.}\label{tab1}
	\scriptsize
	\begin{tabular}{c|c|c|c|c|c|c|c|c}
		\hline
		\multirow{2}*{\textbf{End-to-End}} & \multicolumn{3}{|c|}{\textbf{Cholec80}} & ~ & \multirow{2}*{\textbf{Ours}} & \multicolumn{3}{|c}{\textbf{Cholec80}} \\
		\cline{2-4} \cline{7-9}
		~ & \textbf{Acc} & \textbf{JACC} &\textbf{Rec} & ~ & ~ & \textbf{Acc} & \textbf{JACC} &\textbf{Rec} \\
		\cline{1-4} \cline{6-9}
		\textit{NULL} & 88.8$\pm$6.3 & 73.2$\pm$9.8 & 84.9$\pm$7.2 & ~ & ~& - & - & - \\
		\hline
		
		GRU & 87.1$\pm$7.8 & 69.7$\pm$12.6 & 83.2$\pm$9.4& ~ & GRU& 90.8$\pm$7.0 & 75.5$\pm$11.1 & 85.6$\pm$10.0 \\
		causal TCN & 87.7$\pm$6.3 &77.7$\pm$11.2 &84.3$\pm$6.3& ~ & causal TCN &  91.0$\pm$5.2 & 74.2$\pm$11.8  & 84.1$\pm$9.6 \\
		TCN & 89.8$\pm$6.6 & 75.8$\pm$8.4 & 87.4$\pm$7.5&  ~ & TCN & \textbf{92.8$\pm$5.0} & \textbf{78.7$\pm$9.4}  & \textbf{87.5$\pm$8.3}\\  
		\hline
	\end{tabular}
\end{table}

\begin{figure}
	\centering
	\includegraphics[width=0.9\textwidth]{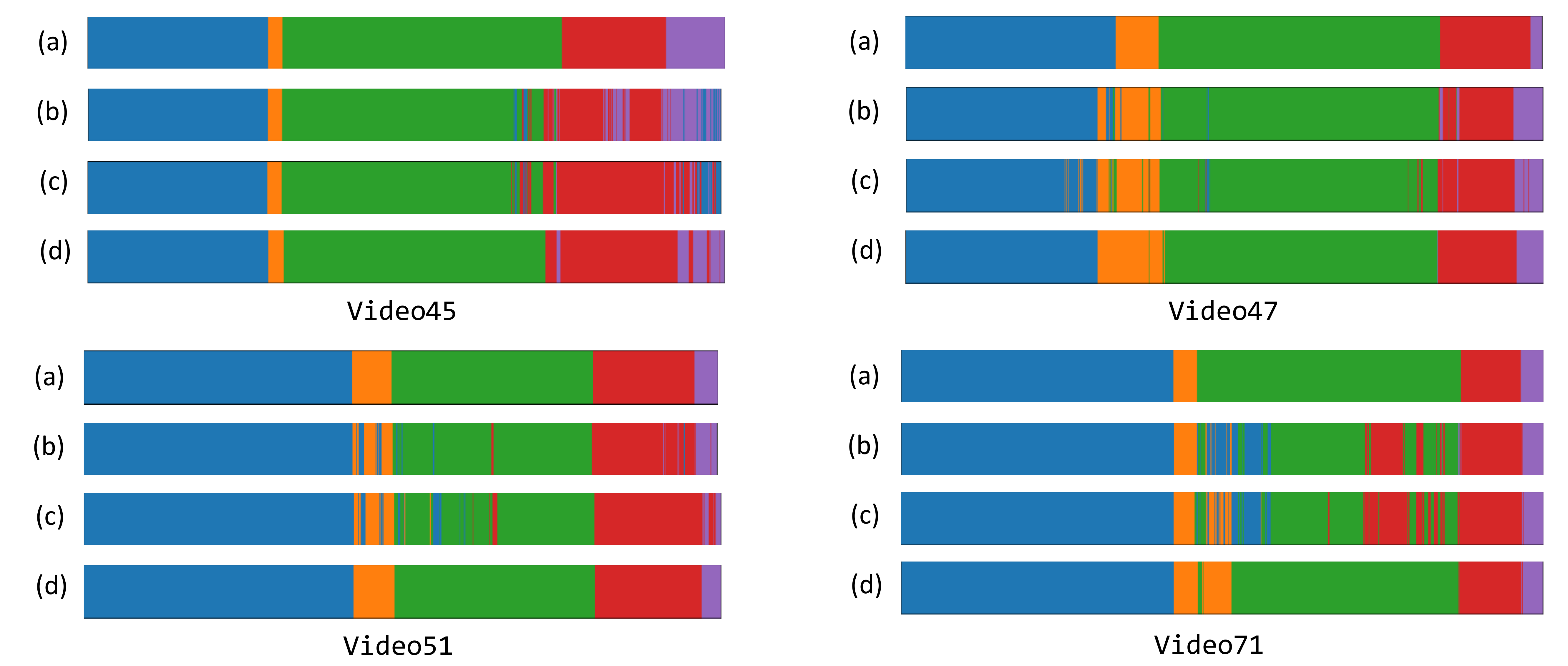}
	\caption{Qualitative Results on Cholec80 dataset of the multi-stage architecture trained with end-to-end manner and ours. Causal TCN and GRU are used for predictor stage and refinement stage, respectively. (a) is the ground truth. (b) is the prediction from the single-stage predictor. (c) is the prediction from multi-stage architecture trained with end-to-end manner. (d) is the prediction from the multi-stage architecture trained with our strategy.} \label{vis}
\end{figure}

\subsection{End-to-End VS. Not End-to-End}
\label{ete}
In this section, we evaluate the performance of multi-stage architecture with the end-to-end training and ours non end-to-end training strategy on the Cholec80 dataset. Results are shown in Table~\ref{tab1}. We can observe that, with the end-to-end training, the multi-stage architecture only achieves comparable results with the single-stage predictor. When we use the causal TCN and GRU as the refinement model, performance is even slightly worse possibly due to the over-fitting problem. Such results are consistent with Czempiel et al.\cite{tecno}. Meanwhile, all three refinement models trained with our proposed disturbed sequence largely boost the performance of the predictor, which shows that our previous analysis above the multi-stage architecture and our solution are not sensitive to the choices of different refinement models. Although TCN achieves the best performance as the refinement model, however, it does not meet the constraint of online. So, we explore the GRU as the refinement model for further experiments. Fig.~\ref{vis} shows qualitative results of GRU as the refinement model under the end-to-end training and ours. The results clearly highlight the ability of multi-stage architectures trained with our solution to obtain consistent and smooth predictions.

\subsection{Stacked Version of Refinement Model}
The refinement model takes an initial prediction as the input, and then outputs a refined prediction. It is natural to come up with the idea that stacks multiple models sequentially so that each model successively operates on the refined predictions of the previous one. In this section, we explore the stacked GRU as the refinement stage model. During the training, the stacked GRU is trained as a whole with the two proposed sequences and loss function is applied on the output of each GRU. Results of stacked GRU with different numbers of GRU are shown in Table~\ref{tab2}. We get the best results when stacking 3 GRUs sequentially on Cholec80 dataset since such composition is an incremental refinement. However, adding more GRUs may cause over-fitting problem due to the increase of parameters. We also conduct experiments for M2CAI16 dataset and get the best results when stracking 2 GRUs, which is less than that of Cholec80 dataset. This may due to the fact that the size of M2CAI16 is smaller.
\begin{table}[t]
	\centering
	\caption{Effects of stacked number of GRUs on the Cholec80 dataset.}\label{tab2}
	\scriptsize
	\begin{tabular}{c|c|c|c|c|c|c}
		\hline
		\multirow{2}*{\textbf{Methods}} & \multicolumn{3}{|c}{\textbf{Cholec80}} & \multicolumn{3}{|c}{\textbf{M2CAI16}}\\
		\cline{2-4} \cline{4-7}
		~ & \textbf{Acc} & \textbf{JACC} & \textbf{Rec} & \textbf{Acc} & \textbf{JACC} & \textbf{Rec} \\
		\hline
		Single GRU         & 90.8$\pm$7.0 & 75.5$\pm$11.1 & 85.6$\pm$10.0 & 86.2$\pm$9.1 & 72.6$\pm$11.6 & 90.0$\pm$11.7\\
		Stacked GRU\tiny{(2 GRU)} & 91.3$\pm$6.1 & 75.9$\pm$11.2 & 84.4$\pm$10.0 & \textbf{88.2$\pm$8.5} & \textbf{75.1$\pm$10.6} & \textbf{91.4$\pm$11.2}\\
		Stacked GRU\tiny{(3 GRU)} & \textbf{91.5$\pm$7.1} & \textbf{77.2$\pm$11.2} & \textbf{86.8$\pm$8.5} & 86.9$\pm$10.2 & 72.7$\pm$11.1 & 89.9$\pm$9.2\\
		Stacked GRU\tiny{(4 GRU)} & 90.8$\pm$5.8 & 74.2$\pm$15.9 & 83.1$\pm$14.6 & 87.0$\pm$8.4 & 72.4$\pm$11.0 & 89.0$\pm$12.3\\
		\hline
	\end{tabular}
\end{table}

\begin{table}[t]
	\centering
	\caption{Results of the multi-stage architecture trained with different disturbed sequence. Abbreviations: \textit{cv} for \textit{cross-validate} type, \textit{mhf} for \textit{mask-hard-frame} type, \textit{rm} for \textit{random-mask} type.}\label{tab3}
	\begin{tabular}{l|c|c|c}
		\hline
		\multirow{2}*{\textbf{Methods}} & \multicolumn{3}{|c}{\textbf{Cholec80}} \\
		\cline{2-4}
		~ & \textbf{Acc} & \textbf{JACC} & \textbf{Rec}\\
		\hline
		cv & 89.6$\pm$5.6 & 70.4$\pm$13.5 & 83.5$\pm$13.2 \\
		mhf& 88.7$\pm$9.4& 70.6$\pm$9.4 & 81.8$\pm$9.9 \\
		rm& 86.5$\pm$7.8 & 69.7$\pm$10.9 & 82.4$\pm$7.0\\
		cv+rm& 90.5$\pm$6.1 & 74.8$\pm$11.8 & \textbf{87.1$\pm$7.0}\\
		cv+rm+mhf & 91.0$\pm$6.8 & 75.3$\pm$13.4 & 85.4$\pm$10.3 \\
		cv+mhf& \textbf{91.5$\pm$7.1} & \textbf{77.2$\pm$11.2} & 86.8$\pm$8.5\\
		\hline
	\end{tabular}
\end{table}

\subsection{Impact of Disturbed Prediction Sequence}
The disturbed prediction sequence for training the refinement stage is very important. In this section, we conduct ablative experiment by using different combinations of the disturbed prediction sequence. For the multi-stage architecture, we use the stacked GRU with 3 GRUs as the refinement stage. Besides the \textit{mask-hard-frame} type and the \textit{cross-validate} type, we additionally designed the \textit{random-mask} type. Different from \textit{mask-hard-frame} type, the \textit{random-mask} type randomly masks frames. Table~\ref{tab3} shows the results on the Cholec80 dataset. First, if we only use one type of disturbed sequence, the performance will drop due to the lack of training data. Meanwhile, we can observe that, the refinement stage trained with \textit{random-mask} type is not as good as \textit{mask-hard-frame} type. This demonstrates the importance of the location to add perturbations.

\subsection{Comparison with the SOTA}
In order to compare our solution with the SOTA methods, we use a stacked GRU with 2 GRUs and a stacked GRU with 3 GRUs as the refinement stage for M2CAI16 dataset and Cholec80 dataset, respectively. Table~\ref{tab4} shows the results. Compared with the single-stage predictor causal TCN, multi-stage architecture trained with our solution largely boost its performance. Noting that TeCNO proposed by Czempiel et al.\cite{tecno} is also a multi-stage network, where causal TCN is used for both predictor stage and refinement stage. We can observe that, when simply applying the multi-stage network on surgical phase recognition task, the end-to-end training makes the refinement ability fall out of its wishes, TeCNO only achieves comparable results with the single-stage predictor causal TCN. The SOTA performance and performance improvement demonstrate the correctness of our analysis and the effectiveness of our solutions.

\begin{table}[t]
	\centering
	\caption{Comparison with the SOTA on Cholec80 and MICAI16 dataset.}
	
	\label{tab4}
	\scriptsize
	\begin{tabular}{c|c|c|c|c|c|c}
		\hline
		\multirow{2}*{\textbf{Methods}} & \multicolumn{3}{|c|}{\textbf{Cholec80}} & \multicolumn{3}{|c}{\textbf{MICAI16}} \\
		\cline{2-4} \cline{5-7}
		~ & \textbf{Acc} & \textbf{JACC} &\textbf{Rec}  & \textbf{Acc} & \textbf{JACC} & \textbf{Rec}\\
		\cline{1-4} \cline{5-7}
		ResNet\cite{hard_frame} & 78.3$\pm$7.7 & 52.2$\pm$15.0 & - & 76.3$\pm$8.9 & 56.4$\pm$10.4 & - \\
		PhaseLSTM\cite{Twinanda2016Single} & 80.7$\pm$12.9 & 64.4$\pm$10.0 & - & 72.5$\pm$10.6 & 54.8$\pm$8.9 & -  \\
		PhaseHMM\cite{Twinanda2016Single} & 71.1$\pm$20.3 & 62.4$\pm$10.4 & -  & 79.5$\pm$12.1 & 64.1$\pm$10.3 &- \\
		EndoNet\cite{Twinanda_endo} & 81.7$\pm$4.2 & - & 79.6$\pm$7.9 & - & - & -  \\  
		{EndoNet-GTbin\cite{Twinanda_endo}} & 81.9$\pm$4.4 & - & 80.0$\pm$6.7 & - & - & -\\
		SV-RCNet\cite{jin2018sv} & 85.3$\pm$7.3 & - & 83.5$\pm$7.5 & 81.7$\pm$8.1 & 65.4$\pm$8.9 & 81.6$\pm$7.2 \\
		OHFM\cite{hard_frame} & 87.0$\pm$6.3 & 66.7$\pm$12.8 & - & 84.8$\pm$8.0 & 68.5$\pm$11.1 & -\\
		TeCNO\cite{tecno} & 88.6 $\pm$ 2.7 & - & 85.2 $\pm$ 10.6 & - & - & -\\ 
		\hline
		causal TCN \tablefootnote{{\color{black} The single-stage predictor.}} & 88.8$\pm$6.3 & 73.2$\pm$9.8 & 84.9$\pm$7.2 & 84.1$\pm$9.6 & 69.8$\pm$10.7 & 88.3$\pm$9.6 \\
		Ours & \textbf{91.5$\pm$7.1} & \textbf{77.2$\pm$11.2} & \textbf{87.2$\pm$8.2} & \textbf{88.2$\pm$8.5} & \textbf{75.1$\pm$10.6} & \textbf{91.4$\pm$11.2} \\ 
		\hline
		
	\end{tabular}
\end{table}

\section{Conclusion and Future Work}
In this paper, we observe that the end-to-end training manner makes the refinement ability of multi-stage architecture fall out of its wishes. In order to solve the problem, we propose a new non end-to-end training strategy and explore different designs of multi-stage architecture for surgical phase recognition task. In the future, we will explore other different predictor models, and apply our solution to other computer vision tasks where the multi-stage architecture is widely applied.

%
%
%
%
\bibliographystyle{splncs04}
\bibliography{reference}
\end{document}